\documentclass[10pt,twocolumn,letterpaper]{article}

\usepackage{sty}
\usepackage{times}
\usepackage{epsfig}
\usepackage{graphicx}
\usepackage{amsmath}
\usepackage{amssymb}
\usepackage{multirow}
\usepackage{booktabs} 
\usepackage{tabularx}
\usepackage{caption}
\usepackage{subcaption}
\newcommand{\sdev}[1]{\textsubscript{{\tiny$\pm#1$}}}
\usepackage{pifont}

    \usepackage{algorithm}
        \usepackage{algorithmic}
        \usepackage{adjustbox}
 \usepackage{xcolor}
\usepackage{url}

\usepackage{csquotes}
\usepackage{float}
\usepackage{bm}

\usepackage{amsfonts}       % blackboard math symbols
\usepackage{nicefrac}       % compact symbols for 1/2, etc.
\usepackage{microtype}      % microtypography

\usepackage{soul}
\usepackage{rotating}

\newcolumntype{?}[0]{!{\vrule width 0.05mm}}

\usepackage{dsfont}

\newcommand\blfootnote[1]{%
  \begingroup
  \renewcommand\thefootnote{}\footnote{#1}%
  \addtocounter{footnote}{-1}%
  \endgroup
}

\usepackage[breaklinks=true,bookmarks=false]{hyperref}

\vcvcfinalcopy % *** Uncomment this line for the final submission

\begin{document}

%%%%%%%%% TITLE
\title{BLADE: Bias-Linked Adaptive DEbiasing}

\author{
Piyush Arora$^{\star}$, \hspace{0.5cm} Navlika Singh$^{\star}$, \hspace{0.5cm} Vasubhya Diwan$^{\star}$, \hspace{0.5cm} Pratik Mazumder$^{\star,\copyright}$\\
Indian Institute of Technology Jodhpur, India \hspace{0.5cm}\\
{\tt\small arora.8@alumni.iitj.ac.in, singh.119@alumni.iitj.ac.in, diwan.1@iitj.ac.in, pratikm@iitj.ac.in}
}

\maketitle

\ifvcvcfinal\thispagestyle{empty}\fi

%%%%%%%%% ABSTRACT
\begin{abstract} 
Neural networks have revolutionized numerous fields, yet they remain vulnerable to a critical flaw: the tendency to learn implicit biases, spurious correlations between certain attributes and target labels in training data. These biases are often more prevalent and easier to learn, causing models to rely on superficial patterns rather than task-relevant features necessary for generalization. Existing methods typically rely on strong assumptions, such as prior knowledge of these biases or access to bias-conflicting samples, i.e., samples that contradict spurious correlations and counterbalance bias-aligned samples, samples that conform to these spurious correlations. However, such assumptions are often impractical in real-world settings. We propose \textbf{BLADE} (\textit{\textbf{B}ias-\textbf{L}inked \textbf{A}daptive \textbf{DE}biasing}), a generative debiasing framework that requires no prior knowledge of bias or bias-conflicting samples. BLADE first trains a generative model to translate images across bias domains while preserving task-relevant features. Then, it adaptively refines each image with its synthetic counterpart based on the image's susceptibility to bias. To encourage robust representations, BLADE aligns an image with its bias-translated synthetic counterpart that shares task-relevant features but differs in bias, while misaligning it with samples sharing the same bias. We evaluate BLADE on multiple benchmark datasets and show that it significantly outperforms state-of-the-art methods. Notably, it exceeds the closest baseline by an absolute margin of around 18\% on the corrupted CIFAR-10 dataset under the worst group setting, establishing a new benchmark in bias mitigation and demonstrating its potential for developing more robust deep learning models without explicit supervision.
\end{abstract}

%%%%%%%%% BODY TEXT
\section{Introduction} 
\blfootnote{$^\star$ All the authors contributed equally.}\blfootnote{$\copyright$ Corresponding author} Neural networks have achieved remarkable success across a wide range of tasks, from image classification \cite{imageclf1, imageclf2} to natural language processing \cite{nlp2020}. However, when trained on real-world data, these models often rely on spurious correlations in the training set rather than task-relevant features \cite{biasproof1, biasproof2}. For example, a model might mistakenly associate snowy backgrounds with skiing, leading to misclassifications in scenarios where such correlations break down. The challenge becomes more severe when bias-conflicting examples are scarce or when explicit bias annotations are unavailable \cite{fightingfire2023}. As a result, models struggle to generalize in settings that deviate from the dominant patterns in the training data, underscoring the need for robust bias mitigation strategies.

Recent approaches to debiasing often rely on prior knowledge of the bias, using group labels or protected attributes to enforce invariance during training \cite{learningnottolern2019, rebias2020, groupdro2020, hybridsample2024}. While effective on controlled benchmarks, these methods face practical limitations due to the high cost and subjectivity associated with acquiring reliable bias annotations. A second category of methods leverages naturally occurring bias-conflicting samples \cite{lff2020, ldd2021, selecmix2022, selfinfluence2024}, typically identified through loss functions or influence-based heuristics. However, the success of these approaches is contingent on the availability of such bias-conflicting samples, which may be sparse or entirely absent in many datasets. Generative methods, such as \cite{biaswap2021, fightingfire2023}, offer a promising alternative by synthesizing bias-conflicting samples. Nonetheless, these approaches often lack mechanisms to calibrate the model’s reliance on generated data, making them vulnerable to performance degradation in settings with strong or entangled biases. This highlights a critical open challenge: \textit{How can we train robust, generalizable models in the absence of both prior bias knowledge and naturally occurring bias-conflicting samples?} 

To address the aforementioned challenges, we propose a generative debiasing framework, \textbf{BLADE} (\textbf{B}ias-\textbf{L}inked \textbf{A}daptive \textbf{DE}biasing), which harnesses the generative capabilities of GANs to translate images across bias domains. Central to BLADE is an \textit{adaptive refinement} mechanism that preserves task-relevant features based on the image’s susceptibility to bias. To encourage robust, bias-invariant representations, BLADE enforces \textit{consistency} between an image and its bias-translated synthetic counterpart, guiding the model to focus on shared, task-relevant cues. Furthermore, it introduces a \textit{contrastive objective} that pushes apart samples from the same bias domain, thereby explicitly discouraging reliance on spurious correlations.

Our key contributions are as follows:
\begin{itemize}
\item We propose \textbf{BLADE}, a novel generative debiasing framework that facilitates bias-invariant representation learning.
\item BLADE functions \emph{without} access to bias-conflicting samples and operates \emph{without requiring any prior knowledge} of the bias type or source, making it broadly applicable across diverse domains and datasets.
\item Unlike existing methods that degrade under fully biased regimes, BLADE exhibits strong generalization even when all training samples contain spurious correlations, demonstrating its robustness in the most challenging learning scenarios.
\end{itemize}

\section{Related Works}
\label{sec:relatedWorks}

Bias mitigation has been extensively studied, with approaches broadly categorized into \ref{sec:withBiasAnnotations} methods requiring explicit bias annotations, \ref{sec:withBiasConfSamples} methods leveraging naturally occurring bias-conflicting samples, and \ref{sec:generative} generative methods synthesizing bias-conflicting samples.

\subsection{Methods using explicit Bias Annotations}  
\label{sec:withBiasAnnotations}
Methods in this category utilize prior information either in the form of known bias annotations or group labels to guide the learning process. For example, \cite{learningnottolern2019, debian2022} introduces a regularization loss based on mutual information between feature embeddings and bias-attributes to discourage the learning of bias-aligned features. \cite{rebias2020} utilizes a loss that penalizes similarity between the de-biased model’s features and the biased model’s features, while distributionally robust optimization techniques, such as \cite{groupdro2020}, optimize performance on the worst-case bias subgroups.

\subsection{Methods utilizing naturally occurring Bias-Conflicting Samples} 
\label{sec:withBiasConfSamples}
These methods identify and leverage naturally occurring bias-conflicting samples. Techniques like \cite{lff2020, jtt2021, debian2022} re-weight samples based on the gap between biased and de-biased models, assuming that samples with higher loss are less biased. \cite{pgi2021} first trains a lightweight reference model to detect spurious background cues, then split each class’s data into majority (bias-aligned) and minority (bias-conflicting) groups. Finally, they add a Predictive Group Invariance loss that aligns the groups’ average softmax outputs via KL‑divergence, forcing the network to rely on robust features rather than shortcuts. \cite{eiil2021} first partitions the data into two pseudo-environments by finding the split that maximizes an invariant-risk penalty on a fixed reference model, and then it trains a new model with an IRM objective across these inferred environments. \cite{selfinfluence2024} leverages influence functions to detect under-fit or mislabeled samples that tend to break spurious correlations. Data augmentation strategies, such as \cite{selecmix2022}, apply mixup to contradicting pairs of examples to promote robust feature learning. \cite{ldd2021} employs disentangled representation learning to synthesize diverse bias-conflicting samples by swapping latent features. \cite{pruning2023} introduces a pruning framework that uses contrastive weight pruning to extract unbiased sub-networks, effectively filtering out weights corresponding to spurious features, whereas \cite{intrinsicfeatures2024} enhances intrinsic feature learning by amplifying the signal from bias-contrastive pairs

\subsection{Methods based on Generative Approaches} 
\label{sec:generative}
Generative methods aim to synthesize bias-conflicting samples or manipulate bias attributes to promote robust representation learning. Early approaches, such as \cite{biaswap2021}, perturb bias regions to break spurious associations. Recent works like \cite{fightingfire2023} employ image translation models to shift bias modes, applying contrastive losses across generated variants to enforce bias-invariance where as \cite{biasadv2023} adversarially perturbs inputs against a spurious‑cue–biased auxiliary model to synthesize bias‑conflicting images, then augments training with these samples so the final network learns robust features rather than shortcuts requiring no bias annotations.

\section{Problem Setting}
\label{sec:problem_setting}

We consider the standard supervised classification setup in which the training dataset consists of samples \( (x_i, y_i) \in \mathcal{X} \times \mathcal{Y} \). Each image \( x_i \) contains both task-relevant attributes \( c_i \in \mathcal{C} \) and potentially spurious or bias-inducing attributes \( b_i \in \mathcal{B} \) where $\mathcal{C, B}$ represents set of task and bias domains across the data. The goal is to learn a model $M = (E,fc)$, where $E$ represents a feature extractor and $fc$ represents a classifier that are parameterized by $\theta$ and $\phi$, respectively. The model $M$ predicts labels \( y_i \) by maximizing the conditional likelihood:

\begin{align}
M_{\theta, \phi}(x_i)
&= \arg\max_{y_i \in \mathcal{Y}} P(y_i \mid \theta, \phi, x_i) \\
&= \arg\max_{y_i \in \mathcal{Y}} P(y_i \mid \theta, \phi, b_i, c_i).
\end{align}

where $\theta$ and $\phi$ denote the model parameters. In biased datasets, strong correlations between \( b_i \) and \( y_i \) during training can cause the model to rely disproportionately on \( b_i \), thus approximating the following,
\begin{align}
\arg\max_{y_i \in \mathcal{Y}} P(y_i \mid \theta, \phi, b_i, c_i) \approx \arg\max_{y_i \in \mathcal{Y}} P(y_i \mid \theta, \phi, b_i),
\end{align}
which leads to weak generalization under distribution shifts where \( b_i \) no longer correlates with the true label.
    
We focus on the particularly challenging \textit{fully biased regime}, previously explored by \cite{fightingfire2023}, in which the training data contains no bias-conflicting samples and no explicit bias annotations are provided.

\section{Method}
\label{sec:Method}
\begin{figure*}[ht]
    \centering
    \includegraphics[width=1.0\linewidth]{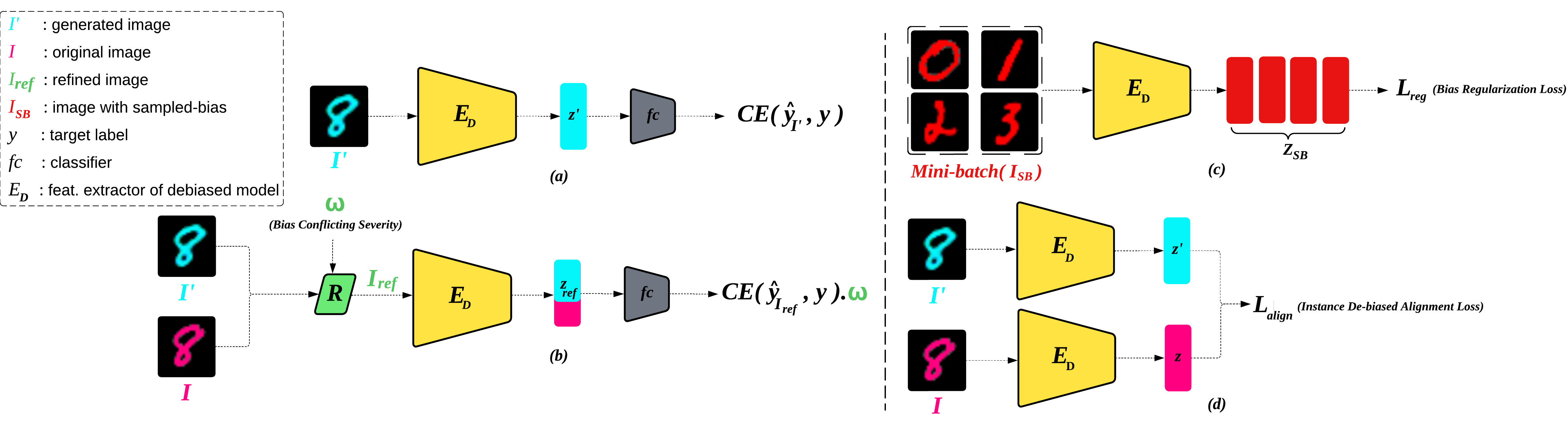}
    \caption{(a) The model is trained on bias-translated generated samples. (b) Refined samples are created by adaptively mixing original and bias-translated generated samples based on bias-conflicting severity (BCS). (c) Bias Regularization loss is computed on samples translated to a common sampled bias domain, promoting invariance across bias-translated sample variants. (d) Instance De-biased Alignment loss is computed that encourages consistency between the original samples and their bias-translated counterparts.
    }
    \label{fig:blade_algo}
\end{figure*}

In this section, we present BLADE, our proposed generative debiasing framework. Section \ref{sec:genbiastrans} introduces the generative model used to transform bias attributes while preserving task-relevant features and Section \ref{sec:clftrans} explains how these counterparts are used directly for training. Section \ref{sec:instlevalign} describes the instance-level alignment mechanism that aligns original and generated samples. In Section \ref{sec:biasinvardiscreg}, we detail a regularization strategy. Section \ref{sec:adaprefine} proposes an adaptive refinement method based on the estimated susceptibility of samples to bias. Finally, Section \ref{sec:trainobj} presents the overall training objective that unifies all components. An overview of our approach is illustrated in Figure \ref{fig:blade_algo}.

\subsection{Generative Model for Bias Translation}\label{sec:genbiastrans} 
Prior works \cite{cyclegan2017, fightingfire2023} have shown that discriminators in generative adversarial networks (GANs) are vulnerable to being influenced by spurious correlations present in biased training data. This entanglement can lead generators to reproduce bias-aligned features rather than performing the intended semantic transformations. Notably, StarGAN \cite{stargan2018} incorporates an auxiliary domain classifier $D_{\text{cls}}$ atop the discriminator $D$, trained to optimize domain classification over real images via the following objective:
\begin{align}
\mathcal{L}_{\text{cls}} = \mathbb{E}_{(x_i, y_i) \sim \mathcal{(X, Y)}}[-\log D_{\text{cls}}(y_i \mid x_i)].
\end{align}
When trained in a biased regime, $D_{\text{cls}}$ often learns to associate domain labels with spurious attributes, causing the generator to translate bias attributes preserving task-relevant features \cite{fightingfire2023}.

We adapt the StarGAN framework~\cite{stargan2018} to train a generator $G$ that translates an image from one bias domain to another while preserving task-relevant features, enabling the construction of bias-translated counterparts for any input. A basic implementation of StarGAN involves the target domain label being spatially replicated and concatenated with the image along the channel dimension. In contrast, we replace this with a learned domain representation obtained via a lightweight encoder, which maps the target image and domain label to a continuous embedding. This encoded representation is then injected into the generator via two AdaIN (Adaptive Instance Normalization) \cite{adain2017} blocks placed before each upsampling layer. Full implementation details and training setup are provided in the Appendix.

\subsection{Classification on Generated Counterparts}\label{sec:clftrans}
Let $M_d$ be our de-biased model, which we are training with the intention of giving de-biased task predictions. Given an input image $x_i$ with label $y_i$, we sample a target label $y_j$ and generate a bias-translated counterpart $x_i^j = G(x_i, y_j)$ such that $y_j \ne y_i$, where $G$ denotes the adapted StarGAN generator from Section \ref{sec:genbiastrans}. We train our de-biased model for the task of classification on these bias-translated generated counterparts using cross entropy loss as:
\begin{align}\label{eq:lossgen}
    \mathcal{L}_{gen} = \mathcal{L}_{CE}(M_d(x_i^j), y_i)
\end{align}
where $\mathcal{L}_{CE}$ represents the Cross Entropy Loss. We apply this loss to expose the model to alternate bias-translated views of the same image as the training progresses, encouraging it to focus on task-relevant features that remain consistent across these views.

\subsection{Instance Level Alignment}\label{sec:instlevalign}
We define an instance-level alignment objective that brings each image closer to its bias-translated counterpart along with  We sample a batch of $N$ samples and construct a paired batch of original and bias-translated counterparts, denoted $\{(x_i, x_i^j)\}_{i=1}^{N}$ where $y_j\ne y_i \forall i$. These are passed through the feature extractor of the de-biased model to obtain the normalized features, i.e., $z_i = E_d(x_i)$ and $z_i^j = E_d(x_i^j)$ for some $x_i$ and $x_i^j$, respectively. To align each sample only with its own generated counterpart, we define a temperature-scaled instance-level objective inspired by contrastive formulations. The Instance De-biased Alignment Loss $\mathcal{L}_{\text{align}}$ over two batched feature representations $Z, Z' \in \mathbb{R}^{N \times d}$ where $Z = \{z_i\}_{i=1}^{N}$ and $Z' = \{z_i^j\}_{i=1}^{N}$ is computed as:
\begin{align}\label{eq:insancede-biasedalign}
\mathcal{L}_{\text{align}}(Z, Z') = - \frac{1}{N} \sum_{i=1}^{N} \log \frac{\exp(z_i^\top z_i^j / \tau)}{\sum_{k=1}^{N} \exp(z_i^\top z_k^j / \tau)},
\end{align}
where $\tau$ is a temperature hyperparameter and all feature vectors are $\ell_2$-normalized prior to computing similarities. In practice, this corresponds to computing a softmax distribution over dot product similarities between original and generated features across the batch.

Our instance-specific application avoids encouraging similarity across all examples sharing the same target label which risks amplifying effects of shared bias attributes, especially when generated samples are imperfect and retain residual spurious cues.

\subsection{Bias-Invariant Discrimination Regularization}\label{sec:biasinvardiscreg}
To further enforce bias-free feature learning, we introduce a regularization objective that explicitly penalizes representational similarity between instances sharing the same bias domain. At each training iteration, we randomly sample a target label and map all images in the batch to the bias domain associated with the target label using the generator $G$. This results in a set of translated samples that all share bias attributes from the same domain.

Let $z_i^{SB}$ be the $\ell_2$-normalized feature representations of an image $x_i$ after translation to the target bias domain associated with class $y_{SB}$. To discourage the model from encoding shared bias-specific features, we define the Bias Regularization Loss $\mathcal{L}_{\text{reg}}$ as a function over the batch of features $Z_{SB} =\{z_i^{SB}\}_{i=1}^{N} \in \mathbb{R}^{N \times d}$ :

\begin{equation}\label{eq:biasregularization}
\mathcal{L}_{\text{reg}}(Z^{SB}) = - \frac{1}{N} \sum_{i=1}^{N} \log \frac{\exp({z_i^{SB}}^\top z_i^{SB} / \tau)}{\sum_{j=1}^{N} \exp({z_i^{SB}}^\top z_j^{SB} / \tau)},
\end{equation}
where $\tau$ is a temperature hyperparameter and all feature vectors are $\ell_2$-normalized. This formulation treats each instance as its own positive pair (diagonal term) and all other bias-aligned instances as negatives, thereby promoting self-similarity while suppressing similarity across the batch.

By minimizing off-diagonal similarities in a bias-homogeneous context, the model is explicitly discouraged from relying on bias-inducing features that are common across translated samples. Instead, it is compelled to extract task-relevant features that persist even under uniform bias translations. This targeted regularization complements our de-biased alignment objective by reducing the risk of feature collapse within biased domains.

\subsection{Adaptive Refinement Strategy}\label{sec:adaprefine}

\subsubsection{Bias Conflicting Severity}\label{sec:bcs}
Along with bias-translated data augmentation, the proposed approach also incorporates a loss re-weighting strategy in order to achieve robust debiasing. The loss re-weighting scheme up-weights the loss for bias-conflicting samples while down-weighting the loss for bias-aligned ones during training. We concurrently train an auxiliary bias-sensitive classification model $M_b$ alongside the primary de-biased classification model $M_d$. The role of $M_b$ is to intentionally capture the spurious correlation in the data, \cite{lff2020}. We use the generalized cross-entropy (GCE) loss \cite{gce2018} for $M_b$, which emphasizes easy-to-classify examples and thus amplifies the influence of the bias-inducing features. In effect, $M_b$ quickly latches onto bias cues (e.g., background or texture) that correlate with class labels, achieving high accuracy on bias-aligned samples but struggling on bias-conflicting ones.

To quantify how strongly a given image $x_i$ is bias-aligned, we define a bias conflicting severity $\omega_i$ based on the prediction losses of $M_b$ and $M_d$ \cite{lff2020}, which is computed as: 
\begin{equation}\label{eq:bcs}
\omega_i = \frac{\mathcal{L}_{CE}(M_b(x_i), y_i)}{\mathcal{L}_{CE}(M_b(x_i), y_i) + \mathcal{L}_{CE}(M_d(x_i), y_i)}
\end{equation}

We compute the cross-entropy loss $\mathcal{L}_{CE}(M(x_i), y_i)$ between the model prediction $M(x_i)$ and the ground-truth label $y_i$ to assess how confidently the bias-sensitive model $M_b$ classifies a given input. A low loss from $M_b$ corresponds to a small $\omega_i$, indicating that the prediction likely relies on bias-aligned features. Conversely, a high loss results in a larger $\omega_i$, suggesting that the sample is bias-conflicting and more valuable for encouraging robust, bias-invariant learning. We incorporate $\omega_i$ into our framework in two key ways: first, to inform data refinement through an adaptive refinement strategy; and second, to reweight the training loss such that bias-conflicting samples receive greater emphasis during optimization.

\subsubsection{Adaptive Refinement}\label{sec:refinement}
In contrast to existing approaches that primarily rely on loss-based reweighting schemes \cite{lff2020}\cite{ldd2021}, we propose to explicitly refine each training image through an adaptive interpolation between its original form and a bias-translated counterpart. Given an input $x_i$ and its generated bias-translated counterpart $x_i^j = G(x_i, y_j)$ such that $y_i \ne y_j$, we compute the bias-conflicting severity measure $\omega_i$ to guide a linear refinement at the representation manifold. This refinement may be applied either directly within image space or on intermediate feature representations derived from the classifier backbone. We define this operation as:
\begin{equation}
x_{i_{ref}} = \mathcal{R}(x_i, x_i^j, \omega_i) = \omega_i x_i + (1 - \omega_i) x_i^j.
\end{equation}

Intuitively, the degree of refinement dynamically adapts to the influence of bias cues within each sample. For images strongly aligned with bias cues (i.e., samples where $\omega_i$ is small), the interpolated result closely resembles the bias-translated version $x_i^j$, effectively mitigating dominant bias-aligned features. Conversely, bias-conflicting samples (where $\omega_i$ is large) remain largely unchanged, preserving their intrinsic unbiased characteristics. By training on a continuous spectrum of these adaptively refined samples, the model gradually learns stable and accurate predictions as spurious bias-related features are incrementally attenuated. This fine-grained refinement provides a more nuanced and effective mitigation strategy compared to the abrupt sample reweighting or discarding techniques commonly employed in prior approaches.

\subsection{Training Objective}\label{sec:trainobj}

\begin{algorithm}[t]
\caption{Debiasing with BLADE}
\label{alg:blade}
\begin{algorithmic}[1]
\REQUIRE Batch $(I, y)$, set of labels $Y$, generator $G$,  
\REQUIRE de-biased model $M_d$ = $(E_d, fc_d)$ w/ params $(\theta_d, \phi_d)$,  
\REQUIRE Bias-sensitive model $M_b$ = $(E_b, fc_b)$ w/ params $(\theta_b, \phi_b)$  
\STATE \textbf{Initialize:} optimizer $O_d$ on $(\theta_d,\phi_d)$, optimizer $O_b$ on $(\theta_b,\phi_b)$  
\WHILE{not converged}
    \STATE \textcolor{gray}{\# Image Generation}
    \STATE \quad $y' = \{y_i' \sim Y \mid y_i' \ne y_i \forall y_i \in y \}$ \textcolor{gray}{\#For translation}
    \STATE \quad $I' = G(I, y')$  \textcolor{gray}{\#Bias Translated Counterparts}
    \STATE \quad $y_{\text{SB}} \sim Y$  \textcolor{gray}{\#Random common‐bias label}
    \STATE \quad $I_{\text{SB}} = G(I, y_{\text{SB}})$  \textcolor{gray}{\#Common‐bias samples}
    \STATE \quad $\omega = \{\omega_i \mid \forall x_i\in I \}$ \textcolor{gray}{\#Bias Conflict Severity}
    \STATE \quad $I_{\text{ref}} = \omega I + (1 - \omega) I'$  \textcolor{gray}{\#Image Refinement}

    \STATE \textcolor{gray}{\# Feature Extraction}
    \STATE \quad $Z = E_d(I),\; Z' = E_d(I')$
    \STATE \quad $Z_{\text{ref}} = E_d(I_{\text{ref}}),\; Z_{\text{SB}} = E_d(I_{\text{SB}})$

    \STATE \textcolor{gray}{\# Logit Computation}
    \STATE \quad $\hat{y}_{I'} = fc_d(Z'),\; \hat{y}_{\text{ref}} = fc_d(Z_{\text{ref}})$

    \STATE \textcolor{gray}{\# Loss Computation for $M_d$}
    \STATE \textcolor{gray}{\#\ \ \ \ \ \ \ \ \ \ \ \ \ \ \ \  $\mathcal{L}_{gen}$ \ \ \ \ \ \ \ \ \ \ \ \ \ \ \ \ \ \ \ \  $\mathcal{L}_{ref}$}
    \STATE \quad $\mathcal{L}_{d} = \mathcal{L}_{CE}(\hat{y}_{I'},y) + \omega\,\mathcal{L}_{CE}(\hat{y}_{\text{ref}},y)$
    \STATE \quad \quad $+\,\mathcal{L}_{\text{align}}(Z, Z') + \mathcal{L}_{\text{reg}}(Z_{\text{SB}})$

    \STATE \textcolor{gray}{\# Compute and apply gradients for $M_d$}
    \STATE \quad $\delta_d = \nabla_{\theta_d,\phi_d}\,\mathcal{L}_{d}$
    \STATE \quad $O_d.\text{update}((\theta_d,\phi_d), \delta_d)$

    \STATE \textcolor{gray}{\# Loss Computation for $M_b$}
    \STATE \quad $\hat{y}_b = fc_b(E_b(I))$
    \STATE \quad $\mathcal{L}_{b} = GCE(\hat{y}_b, y)$

    \STATE \textcolor{gray}{\# Compute and apply gradients for $M_b$}
    \STATE \quad $\delta_b = \nabla_{\theta_b,\phi_b}\,\mathcal{L}_{b}$
    \STATE \quad $O_b.\text{update}((\theta_b,\phi_b), \delta_b)$
\ENDWHILE
\end{algorithmic}
\end{algorithm}

We begin by training a StarGAN model on the biased dataset to obtain a generator \( G \) capable of translating bias-inducing attributes while preserving class semantics (see Appendix for details). As outlined in Algorithm~\ref{alg:blade}, we use this generator to synthesize bias-translated counterparts \( I' \), instances translated to a common sampled bias \( I_{\text{SB}} \), and adaptively refined samples \( I_{\text{ref}} \) for each input \( I \). These are leveraged to compute four complementary losses:
\\
\begin{itemize}
    \item \( \mathcal{L}_{\text{gen}} \): Classification loss on bias-translated counterparts as \ref{eq:lossgen},
    \item \( \mathcal{L}_{\text{ref}} \): Weighted classification loss on adaptively refined inputs as \ref{eq:bcs},
    \item \( \mathcal{L}_{\text{align}} \): Instance De-biased Alignment loss between original samples and bias translated counterparts computed as \ref{eq:insancede-biasedalign},
    \item \( \mathcal{L}_{\text{reg}} \): Bias Regularization loss across bias-invariant features of samples translated to a common Sampled Bias domain computed as \ref{eq:biasregularization}
\end{itemize}

The total loss for updating the de-biased model \( M_d \) is:
\begin{equation}
\mathcal{L}_{d} = \mathcal{L}_{\text{gen}} + \mathcal{L}_{\text{ref}} + \mathcal{L}_{\text{align}} + \mathcal{L}_{\text{reg}}.
\end{equation}
In parallel, we optimize the bias-sensitive model \( M_b \) using a generalized cross-entropy objective:
\begin{equation}
\mathcal{L}_{b} = GCE(M_b(I), y),
\end{equation}
which encourages the bias-sensitive model to capture bias-aligned patterns. Together, these losses form the foundation of BLADE’s training pipeline, with each component contributing to robust, bias-invariant representation learning under fully biased supervision.

\section{Experiments}
\label{sec:Experiments}

\begin{table*}[ht]
\centering
\renewcommand{\arraystretch}{1.5}
\setlength{\tabcolsep}{5pt}
\begin{adjustbox}{width=1.00\textwidth,center}

{
\begin{tabular}{c | ccccc | ccccc | cc}
\toprule
\multirow{2}{*}{\textbf{Method}} 
& \multicolumn{5}{c|}{\textbf{Colored MNIST}} 
& \multicolumn{5}{c|}{\textbf{Corrupted CIFAR-10}} 
& \multicolumn{2}{c}{\textbf{bFFHQ}} \\
\cmidrule{2-13}
& 0.0\% & 0.5\% & 1.0\% & 2.0\% & 5.0\%
  & 0.0\% & 0.5\% & 1.0\% & 2.0\% & 5.0\%
  & 0.0\% & 0.5\% \\
\midrule
Vanilla
  & 12.32\sdev{0.91} & 39.00\sdev{0.85} & 56.10\sdev{1.80} & 69.30\sdev{0.25} & 83.90\sdev{0.88}
  & 16.05\sdev{0.13} & 20.87\sdev{0.34} & 24.05\sdev{0.61} & 29.47\sdev{0.20} & 41.12\sdev{0.16}
  & 37.93\sdev{1.88} & 52.40\sdev{1.88} \\
ReBias$^\textsuperscript{†}$
  & 14.60\sdev{0.60} & 70.50\sdev{0.95} & 88.50\sdev{0.20} & 92.80\sdev{0.40} & 96.96\sdev{0.15}
  & 21.93\sdev{0.37} & 22.27\sdev{0.41} & 25.72\sdev{0.20} & 31.66\sdev{0.43} & 43.43\sdev{0.41}
  & 43.07\sdev{0.19} & 59.46\sdev{0.92} \\
LfF 
  & 13.05\sdev{0.60} & 66.20\sdev{1.03} & 79.70\sdev{0.66} & 85.30\sdev{0.10} & 84.30\sdev{1.00}
  & 15.88\sdev{0.45} & 28.58\sdev{0.23} & 33.68\sdev{0.50} & 39.96\sdev{1.09} & 49.49\sdev{0.16}
  & 39.45\sdev{0.07} & 62.20\sdev{0.52} \\
DisEnt

  & 11.12\sdev{0.05} & 65.22\sdev{1.15} & 81.73\sdev{0.45} & 84.79\sdev{0.25} & 89.66\sdev{1.06}
  & 18.76\sdev{0.88} & 28.62\sdev{1.74} & 36.49\sdev{0.03} & 41.51\sdev{2.34} & 51.13\sdev{0.62}
  & 38.06\sdev{1.18} & 63.83\sdev{0.92} \\
Debian 
  & 14.13\sdev{0.25} & 68.37\sdev{1.72} & 82.78\sdev{1.11} & 85.29\sdev{0.29} & 89.40\sdev{1.55}
  & 19.82\sdev{0.61} & 27.98\sdev{1.17} & 38.38\sdev{0.83} & 43.16\sdev{1.93} & 54.50\sdev{2.51}
  & 39.26\sdev{0.43} & 62.80\sdev{0.62} \\
BCSI+SelecMix
  & -- & 87.88\sdev{0.57} & 95.62\sdev{0.35} & \underline{97.15}\sdev{0.30} & \underline{98.13}\sdev{0.01}
  & -- & \underline{39.35}\sdev{0.36} & \underline{45.59}\sdev{0.33} & \underline{53.42}\sdev{0.40} & \underline{58.91}\sdev{0.31}
  & -- & \underline{65.88}\sdev{0.03} \\
CDvG+LfF
  & \underline{96.48}\sdev{0.19} & \underline{96.20}\sdev{0.12} & \underline{96.45}\sdev{0.20} & 96.95\sdev{0.18} & 96.95\sdev{0.06}
  & \underline{29.24}\sdev{0.47} & 38.42\sdev{0.27} & 43.18\sdev{0.35} & 46.83\sdev{0.07} & 51.74\sdev{0.13}
  & \underline{49.57}\sdev{0.48} & 62.16\sdev{0.45} \\
\textbf{BLADE (Ours)}
  & \textbf{96.92}\sdev{0.07} & \textbf{97.11}\sdev{0.19} & \textbf{97.27}\sdev{0.07} & \textbf{97.51}\sdev{0.10} & \textbf{98.34}\sdev{0.17}
  & \textbf{48.18}\sdev{0.07} & \textbf{51.64}\sdev{0.34} & \textbf{54.63}\sdev{0.27} & \textbf{57.76}\sdev{0.25} & \textbf{62.05}\sdev{0.17}
  & \textbf{55.56}\sdev{0.73} & \textbf{71.12}\sdev{1.48} \\
\bottomrule
\end{tabular}
}
\end{adjustbox}

\caption{Accuracy evaluated on the unbiased test datasets for Colored MNIST, Corrupted CIFAR-10, and bFFHQ dataset with varying Bias Conflicting Ratio (BCR). We denote the best results in \textbf{bold} and the second best by \underline{underlining} them. \textsuperscript{†} indicates that the method requires some prior information about the bias. All the reported results are averaged over 3 runs.}
\label{tab:blade_results1}
\end{table*}

\begin{table*}[ht]
\centering
\renewcommand{\arraystretch}{1.5}
\setlength{\tabcolsep}{5pt}
\begin{adjustbox}{width=0.8\textwidth,center}
\begin{tabular}{c | c c c c c c |c c c c c c}
\toprule
\multirow{2}{*}{\textbf{Test type}} & \multicolumn{6}{c|}{\textbf{ResNet-18}} & \multicolumn{6}{c}{\textbf{ResNet-50}} \\
                   & Vanilla & LfF & DisEnt & BPA & CDvG+LfF & \textbf{Ours} & Vanilla & EIIL & JTT & BCSI+SelecMix & CDvG+LfF & \textbf{Ours}\\ 
\midrule
Unbiased           & 84.63 & 85.48 & 83.72  & \underline{87.05} & 86.25 & \textbf{89.29} & \textbf{97.30} & \underline{96.90} & 93.60 & -- &91.30 & 91.74 \\ 
Worst‐group        & 62.39 & 68.02 & 69.54 & 71.39 & \underline{74.92} & \textbf{88.26} & 60.30 & 78.70 & 86.00 & 87.25 & \underline{84.80} & \textbf{91.83} \\ 
\bottomrule
\end{tabular}
\end{adjustbox}
% \caption{Experimental results on the Waterbirds dataset (average accuracy over 3 runs).}
\caption{Accuracy evaluated on the unbiased test set of Waterbirds. We denote the best results in \textbf{bold} and the second best by \underline{underlining} them. Results are reported for ResNet-18 and ResNet-50 after averaging over 3 runs.}
\label{tab:blade_results_wbirds}
\end{table*}

\begin{table*}[ht]
\centering
\renewcommand{\arraystretch}{1.5}
\setlength{\tabcolsep}{5pt}
\begin{adjustbox}{width=0.7\textwidth,center}
\resizebox{\textwidth}{!}{%
\begin{tabular}{c | c | c c c c c c}
\toprule
\textbf{Left Color} 
  & \textbf{Right Color} 
  & \multirow{2}{*}{Vanilla} 
  & \multirow{2}{*}{LfF} 
  & \multirow{2}{*}{EIIL} 
  & \multirow{2}{*}{PGI} 
  & \multirow{2}{*}{DebiAN} 
  & \multirow{2}{*}{\textbf{Ours}} \\ 
\textbf{BCR = 1.0\%} 
  & \textbf{BCR = 5.0\%} 
  & & & & & & \\ 
\midrule
Bias-Aligned     & Bias-Aligned     & 100.00\sdev{0.0} &  99.60\sdev{0.5} & 100.00\sdev{0.0} &  98.60\sdev{2.3} & 100.00\sdev{0.0} &  90.50\sdev{1.3} \\
Bias-Aligned     & Bias-Conflicting &  97.10\sdev{0.5} &   04.70\sdev{0.5} &  97.20\sdev{1.5} &  82.60\sdev{19.6}&  95.60\sdev{0.8} &  93.10\sdev{0.9} \\
Bias-Conflicting & Bias-Aligned     &  27.50\sdev{3.6} &  98.60\sdev{0.4} &  70.80\sdev{4.9} &  26.60\sdev{5.5} &  76.50\sdev{0.7} &  92.36\sdev{0.4} \\
Bias-Conflicting & Bias-Conflicting &   05.20\sdev{0.4} &   05.10\sdev{0.4} &  10.90\sdev{0.8} &   09.50\sdev{3.2} &  16.00\sdev{1.8} &  92.31\sdev{0.2} \\
\midrule
\multicolumn{2}{c|}{\textbf{Overall}}
                 &  57.40\sdev{0.7} &  52.00\sdev{0.1} &  69.70\sdev{1.0} &  54.30\sdev{4.0} &  \underline{72.00}\sdev{0.8} &  \textbf{92.35}\sdev{0.6} \\
\bottomrule
\end{tabular}%
}
\end{adjustbox}
\caption{Accuracy evaluated on multiple configurations of the unbiased test set of Multi-Colored MNIST. We denote the best results in \textbf{bold} and the second best by \underline{underlining} them. Results are reported after averaging over 3 runs.}
\label{tab:blade_results_mcmnist}
\end{table*}

\subsection{Datasets}\label{sec:dataset}
We evaluate our method on two categories of benchmark datasets: synthetically biased datasets, Colored MNIST \cite{lff2020}, Multi-Colored MNIST \cite{debian2022}, and Corrupted CIFAR-10 \cite{lff2020}, and real-world datasets exhibiting naturally occurring bias,  bFFHQ \cite{ldd2021} and Waterbirds \cite{groupdro2020}. Following established protocols, we vary the proportion of bias-conflicting samples from 0.5\% to 5\%, a metric we refer to as the Bias-Conflicting Ratio (BCR), to simulate increasingly biased training conditions. To further assess robustness, we include an extreme setting with 0\% bias-conflicting samples, termed the \textit{fully biased regime}, by removing all unbiased samples from the 0.5\% configuration. Additionally, the Multi-Colored MNIST dataset introduces multiple concurrent biases, providing a more challenging setup; we evaluate our method across these variations and report detailed results accordingly. Implementation details are provided in the Appendix.

\subsection{Baselines}\label{sec:baseline}
We compare BLADE with a range of existing debiasing approaches. ReBias~\cite{rebias2020} uses explicit bias annotations to disentangle and suppress bias-related features during training. LfF~\cite{lff2020} avoids the need for bias labels by training an auxiliary network to capture spurious patterns and upweighting samples that the biased network fails on. DisEnt \cite{disentangledfeature2021} builds on LfF by disentangling semantic and bias features to generate bias-conflicting augmentations, further improving robustness. BCSI~\cite{selfinfluence2024} combined with SelecMix \cite{selecmix2022} leverages self-influence scores to identify bias-conflicting instances and applies label-consistent sample mixing across bias domains. CDvG integrated with LfF \cite{fightingfire2023} uses a generative image translation module to create contrastive views with shifted bias, enhancing LfF’s ability to learn bias-invariant representations even without bias annotations. We also compare our method with JTT \cite{jtt2021}, EIIL \cite{eiil2021}, Debian \cite{debian2022} and PGI \cite{pgi2021}.

\subsection{Evaluation Protocol}\label{sec:evalproto}
Following established protocols, we report the mean accuracy and standard deviation across three independent runs. For Colored MNIST and Corrupted CIFAR-10, we evaluate performance on the respective unbiased test sets. For Multi-Colored MNIST, we present results across multiple configurations of the unbiased test set. On bFFHQ, we report the minority-group accuracy, while for Waterbirds, we report both the worst-group and unbiased accuracies.

\section{Results}
\label{sec:results}

\subsection{Synthetic Datasets}
\label{sec:syntheticData}
Table~\ref{tab:blade_results1} presents the experimental results on the unbiased test sets of the Colored MNIST and Corrupted CIFAR-10 datasets across varying levels of BCR. We observe that BLADE consistently achieves the highest accuracy across all bias settings, including and especially the \textit{fully biased regime}, where existing methods often struggle due to overfitting to spurious cues.

For Colored MNIST, while ReBias performs well at higher BCR levels due to its access to bias annotations, its performance drops significantly at 0\% BCR. LfF and DisEnt, which do not rely on bias labels, show better generalization at moderate BCR but underperform in the \textit{fully biased regime}. CDvG + LfF remains competitive by leveraging synthetic samples, yet falls short of BLADE, which surpasses it (e.g., 96.92\% vs.\ 96.48\% at 0\% BCR). Notably, BCSI + SelecMix, despite strong gains at higher BCR levels, is not applicable when no bias-conflicting samples are available, as it relies on naturally occurring bias-conflicting samples.

For Corrupted CIFAR-10, the relative difficulty of the task amplifies the differences among methods. While the best-performing baselines exhibit a noticeable drop in performance as the BCR decreases, BLADE maintains a substantial margin across all settings. For example, at 0\% BCR, BLADE outperforms the next best method (CDvG + LfF) by over 18\% (48.18\% vs.\ 29.24\%), highlighting its ability to learn task-relevant features even under fully biased settings. This trend persists up to 5\% BCR, where BLADE continues to outperform all baselines.
\subsection{Real-World Datasets}
\label{sec:realData}
The results on the bFFHQ dataset are reported in Table~\ref{tab:blade_results1}. The results indicate that most of the methods see significant performance degradation under highly biased training distributions. While ReBias and DisEnt offer moderate gains, they either rely on bias labels or struggle in the low-BCR regime, respectively. CDvG+LfF achieve a better balance across settings, but BLADE surpasses all baselines by a substantial margin. In the \textit{fully biased regime}, BLADE improves minority-group accuracy by over 6\% compared to the strongest competing method, indicating strong robustness to spurious gender cues despite the absence of bias-conflicting supervision.

The experimental results on the Waterbirds datasets using the ResNet-18 and ResNet-50 architectures are reported in Table~\ref{tab:blade_results_wbirds}. The results indicate that BLADE achieves the highest unbiased accuracy (89.29\%) and substantially outperforms all baselines in worst-group accuracy (88.26\%), improving over the second-best method (CDvG+LfF) by more than 13\%. This large gain in worst-group performance highlights the ability of BLADE to maintain task-consistent decision boundaries even when the spurious context is misaligned with the label. Furthermore, with ResNet-50, BLADE maintains its superiority over prior work, demonstrating its scalability across model capacities.
\subsection{Multi-Bias Setting}
\label{sec:multiBias}
We evaluate BLADE on the Multi-Colored MNIST benchmark \cite{debian2022}, which introduces two independent bias attributes, left and right digit colors, to test debiasing in multi-bias scenarios. Table~\ref{tab:blade_results_mcmnist} reports the experimental results for the four possible combinations of bias-aligned and bias-conflicting color assignments, along with overall unbiased accuracy.
First, existing methods show inconsistent performance across different bias configurations. Vanilla training nearly fails on conflicting-color cases, while LfF and EIIL succeed primarily on the more salient left bias but collapse on conflicting instances. PGI and DebiAN exhibit more balanced performance, yet their overall accuracy remains limited (below 73\%). In contrast, BLADE achieves strong performance across all four combinations, reaching an overall accuracy of 92.35\%, which is over 20\% higher than the second-best method (DebiAN).
Importantly, BLADE maintains high accuracy even when both bias attributes conflict, an especially challenging scenario where most prior methods drop below 20\%, yet BLADE achieves over 92\%.

\section{Ablation Study}
\label{sec:ablation_study}

\subsection{Component-wise Contribution}
\label{sec:component}

\begin{table}[t]
\centering
\setlength{\tabcolsep}{4.5pt}
\renewcommand{\arraystretch}{1.3}
\begin{adjustbox}{width=0.38\textwidth}
\begin{tabular}{l c c c c c}
\toprule
\textbf{Dataset} & \textbf{BCR} & \textbf{BTC} & \textbf{+ AR} & \textbf{+ IAL} & \textbf{+ BRL} \\
\midrule
\multirow{3}{*}{\shortstack{Corrupted\\CIFAR-10}} 

& 0.5\% & 45.71 & 49.06($\uparrow$ 3.35) & 49.93($\uparrow$ 0.87) & 51.64($\uparrow$ 1.71) \\
& 1.0\% & 46.25 & 52.42($\uparrow$ 6.17) & 53.28($\uparrow$ 0.86) & 54.63($\uparrow$ 1.35) \\
& 2.0\% & 47.66 & 55.71($\uparrow$ 8.05) & 56.74($\uparrow$ 1.03) & 57.76($\uparrow$ 1.02) \\

\bottomrule
\end{tabular}
\end{adjustbox}
\caption{Ablation study on Corrupted CIFAR-10 across varying Bias-Conflicting Ratios (BCR). Each row reports accuracy when progressively adding components to the base setup.}
\label{tab:blade_ablation}
\end{table}
\noindent
To assess the contribution of each component in BLADE, we conducted an ablation study (Table \ref{tab:blade_ablation}) using Corrupted CIFAR-10 across varying BCR. The baseline configuration involves training solely on \textit{bias-translated counterparts} (BTC), without any refinement or auxiliary loss. While this setup yields stable accuracy across BCR levels (maintaining around 45.71\% to 47.66\% from 0.5\% to 2.0\% BCR), its performance remains limited. This is because, although the generator alters the bias attribute, it inadvertently suppresses naturally occurring bias-conflicting cues that are essential for robust learning.

Introducing \textit{Adaptive Refinement} (AR) results in consistent improvements of 3.35\% to 8.05\% across BCR levels. By selectively preserving reliable bias-conflicting samples, this component enables the model to retain informative signals otherwise lost through imperfect translation. Notably, the improvement is more pronounced at higher BCRs, where conflicting samples are more prevalent. Further gains are observed upon adding the \textit{Instance De-biased Alignment Loss} (IAL), which provides an additional performance boost 0.87\% to 1.03\% across BCR. This loss encourages consistency in feature representations between original and translated views, reinforcing the model's ability to generalize across domains. Lastly, incorporating \textit{Bias Regularization Loss} (BRL) enhances performance by another 1.71\%-1.02\% by explicitly discouraging shared spurious features among samples with same bias domain. 

Overall, the full BLADE model, combining all three components, consistently achieves the best performance with margins of up to $+10\%$ over its baseline. These results confirm the complementary effect of each module in achieving bias-invariant generalization.

\subsection{Refinement Strategies Comparison}
\label{sec:refineCompare}
\begin{table}[t]
\centering
\setlength{\tabcolsep}{6pt}
\renewcommand{\arraystretch}{1.3}
\begin{adjustbox}{width=0.45\textwidth}
\begin{tabular}{l l c}
\toprule
\textbf{Dataset} & \textbf{Refinement Strategy} & \textbf{Accuracy (\%)} \\
\midrule
\multirow{3}{*}{\shortstack{Corrupted\\CIFAR-10 (2\% BCR)}} 
& Original Images Only & 50.42 \\
& Random Mixup & 53.31 \\
& \textbf{Ours (Adaptive Refinement)} & \textbf{57.76}\\
\bottomrule
\end{tabular}
\end{adjustbox}
\caption{Comparison of interpolation strategies on a biased variant of Corrupted CIFAR-10 (2\% BCR).}
\label{tab:refinement_comparison}
\end{table}

\noindent Table~\ref{tab:refinement_comparison} presents a comparison of refinement strategies on the Corrupted CIFAR-10 dataset with 2\% BCR. Training solely on original images yields an accuracy of 50.42\%, as it reinforces existing bias and fails to leverage informative bias-conflicting instances. Applying random mixup improves performance to 53.3\% by interpolating between samples, but it does not explicitly preserve naturally occurring bias-conflicting examples that are crucial for robust learning. In contrast, our adaptive refinement strategy achieves a significantly higher accuracy of 57.76\% by selectively retaining high-quality bias-conflicting samples. These results highlight the limitations of naive interpolation and demonstrate the effectiveness of informed, instance-level refinement in mitigating dataset bias.

\subsection{Visualizations}
\label{sec:visualize}
\begin{figure}[t]
    \centering
    \includegraphics[width=0.75\linewidth]{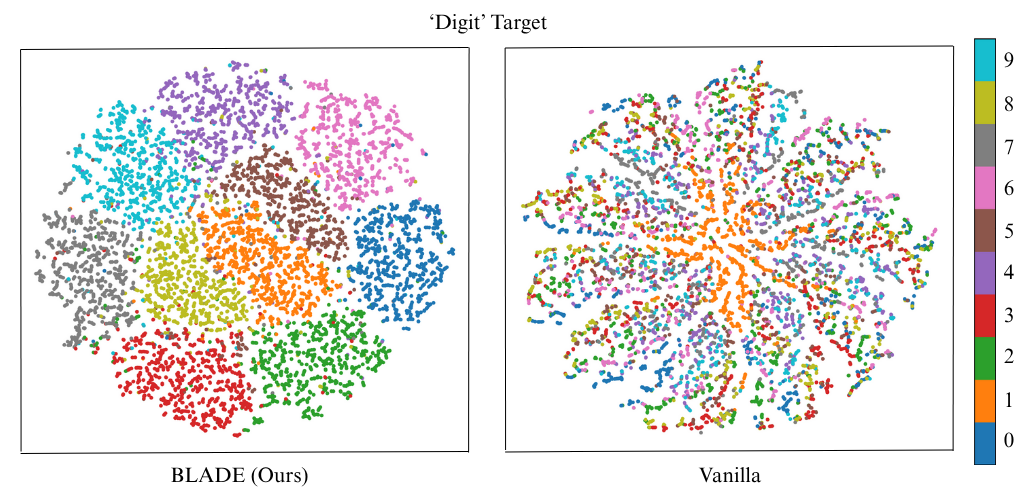}
    \caption{t-SNE visualization comparing feature representations learned by BLADE and the vanilla model on the unbiased test set of Colored MNIST.}
    \label{fig:tsne}
\end{figure}
\begin{figure}[t]
    \centering
    \includegraphics[width=0.50\linewidth]{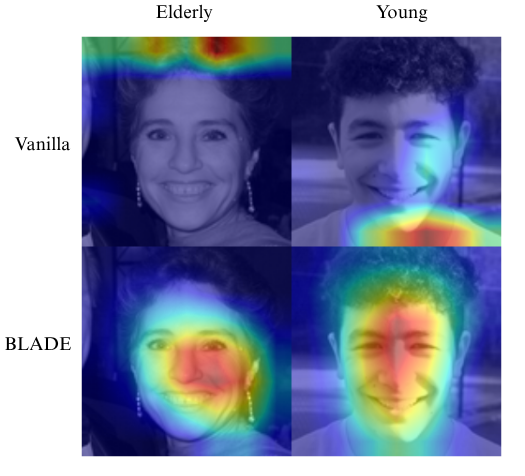}
    \caption{Grad-CAM heatmap visualizations on bFFHQ unbiased test set, comparing the Vanilla model and BLADE.}
    \label{fig:gradcam}
\end{figure}
\noindent To qualitatively assess BLADE’s effect on learned representations, we present t-SNE and Grad-CAM visualizations. On Colored MNIST (Figure~\ref{fig:tsne}), the vanilla model shows poor class separability, while BLADE promotes clustering by digit identity, indicating better task-relevant feature extraction. On bFFHQ (Figure~\ref{fig:gradcam}), Grad-CAM highlights show that the vanilla model focuses on spurious cues like hair and facial contours, reflecting training bias. In contrast, BLADE attends to more semantically meaningful facial regions, aligning with the age prediction task. These results illustrate BLADE’s ability to enhance both performance and interpretability by fostering causally grounded representations.

\section{Conclusion}
In this paper, we introduced \textbf{BLADE}, a generative debiasing framework that operates without the need for bias-conflicting samples or explicit bias annotations. By combining generative translations, adaptive refinement, and feature-level regularization, BLADE facilitates the learning of task-relevant representations even under severely biased training distributions. Through comprehensive experiments on both synthetic and real-world datasets, BLADE consistently outperforms existing approaches, especially in the most challenging scenarios where conventional methods struggle. These results underscore the effectiveness of principled data refinement and representation alignment in achieving robust generalization under extreme bias.

{\small
\bibliographystyle{ieee_fullname}
\bibliography{egbib}
}

{\LARGE \textbf{Appendix}}

\section{Implementation Details}
 
\subsection{Classification Network}
We use a 3-layered Multi-layer Perceptron for Colored MNIST and Multi Colored CMNIST, while ResNet-18 \cite{resnet2015} is employed for Corrupted CIFAR-10, BFFHQ, and WBIRDS. Pretrained ImageNet weights are utilized for WBIRDS, whereas all other datasets are trained from scratch.

The Adam optimizer is used for all datasets. The learning rate is set to 1e-3 for Colored MNIST, Multi Colored MNIST, and Corrupted CIFAR-10, while for BFFHQ and WBIRDS, it is reduced to 1e-4. A cosine learning rate scheduler is applied, gradually decaying the learning rate to 10\% of its initial value throughout training. Training is conducted for 150 epochs on Colored MNIST, Multi Colored MNIST, 100 epochs on BFFHQ, 200 epochs on WBIRDS, and 350 epochs on Corrupted CIFAR-10.

\subsection{StarGAN Adaptation}

\begin{figure}[htbp]
  \centering
  \begin{subfigure}[c]{0.18\textwidth}
    \includegraphics[width=\linewidth]{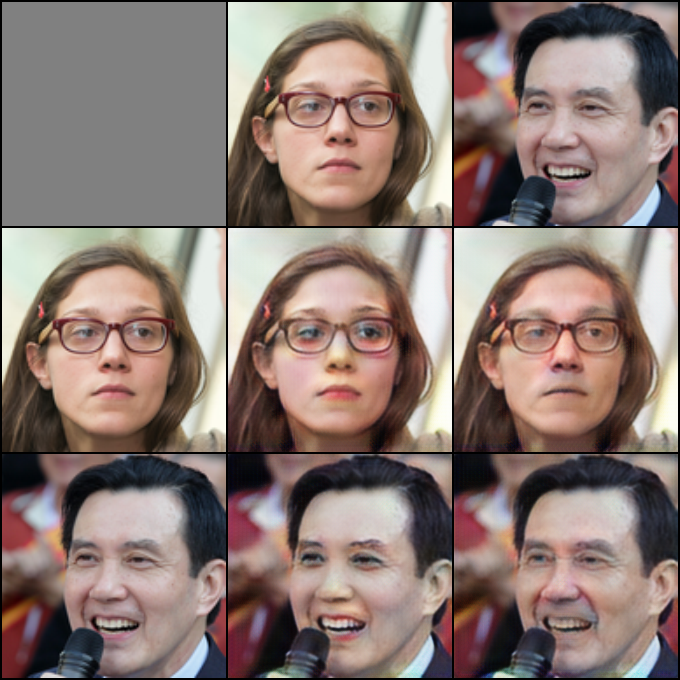}
    \caption{bFFHQ Dataset}
  \end{subfigure}
  \hfill
  \begin{subfigure}[c]{0.28\textwidth}
    \includegraphics[width=\linewidth]{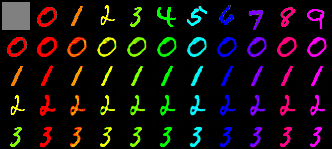}
    \caption{Colored MNIST Dataset}
  \end{subfigure}
  \caption{The figure showcases the ability of the modified StarGAN to translate images across bias domains. The first column represents original instances that are to be translated. The top row represents instances and their bias domain to which the original instances are translated into.}
  \label{fig:two_images}
\end{figure}

\begin{table}
\label{table:stargan}
\centering
\resizebox{0.45\textwidth}{!}{%
\begin{tabular}{cccccc}
\toprule
Dataset & BCR (\%) & \begin{tabular}[c]{@{}c@{}}CDvG\\w/ StarGAN\end{tabular} & \begin{tabular}[c]{@{}c@{}}CDvG\\w/ StarGANv2\end{tabular} & \begin{tabular}[c]{@{}c@{}}BLADE\\w/ StarGAN\end{tabular} & \begin{tabular}[c]{@{}c@{}}BLADE\\ \end{tabular} \\
\midrule
\multirow{2}{*}{\begin{tabular}[c]{@{}c@{}}Corrupted\\CIFAR-10\end{tabular}} 
& 0.0 & 31.50 & 32.30 & 43.34 & 48.18 \\
& 5.0 & 42.75 & 47.80 & 58.54 & 62.05 \\
\bottomrule
\end{tabular}
}
\caption{Ablation study to compare performances of BLADE and CDvG with different GANS}
\end{table}

\subsubsection{Hyperparameters and Configuration}
We implement biased StarGAN \cite{stargan2018} training by adopting the default architectures, optimizers, loss-term weights, and other settings from the official StarGAN  \href{https://github.com/yunjey/stargan}{repository} across all datasets. Our generator uses the basic residual-block design with skip connections: three blocks (each containing two convolutional layers) for Colored MNIST and six blocks for the remaining datasets. The discriminator follows the standard StarGAN structure, with four blocks for Colored MNIST and BFFHQ and five blocks for Corrupted CIFAR-10 and Waterbirds. Unless specified otherwise, we preserve the repository’s default optimizer parameters and loss weightings. For Colored MNIST specifically, we assign a reconstruction-loss weight of 500.
\subsubsection{Learnable Domain-Conditioned Generator}

We adapt the original StarGAN generator to enable flexible and learnable domain translation suitable for bias mitigation. Unlike the original formulation, which conditions the generator by concatenating a spatially replicated one-hot domain label to the input image, our design replaces this hard-coded mechanism with a more expressive, data-driven alternative.

Specifically, we introduce an auxiliary encoder that extracts a domain representation from a reference image corresponding to the target bias domain. This encoder is jointly trained with the generator and outputs a continuous domain feature vector used to guide translation. The resulting embedding captures domain-specific variations and is injected into the generator via Adaptive Instance Normalization (AdaIN)~\cite{adain2017} blocks.

\subsubsection{Encoder Architecture}
The encoder follows a ResNet-style architecture composed of an initial convolutional layer, followed by two downsampling blocks and three residual blocks. These stages progressively extract semantic features while reducing spatial resolution. The output is passed through a convolutional head and global average pooling to produce a compact representation. Finally, a fully connected layer projects this feature into a bank of per-domain style vectors. During inference, we index this bank using the target domain label.

\subsubsection{AdaIN Block}
Each AdaIN block injects domain information into the generator’s feature stream using the encoded style representation. The block consists of two convolutional layers, each followed by AdaIN normalization, with a ReLU activation applied between them. Formally, given an input feature map $x$ and domain code $z$, the AdaIN block is defined as:
\begin{align*}
y_1 &= \text{ReLU}(\text{AdaIN}(\text{Conv}_1(x), z)) \\
y_2 &= \text{AdaIN}(\text{Conv}_2(y_1), z)
\end{align*}
Here, AdaIN normalizes each channel of $x$ using instance normalization and then applies a channel-wise affine transformation conditioned on $z$. The transformation parameters are computed via learned linear projections of the style vector. This enables smooth, content-preserving modulation of the bias domain at multiple layers of the generator.

\subsubsection{Key Differences from StarGAN}

Our generator introduces a learned domain conditioning mechanism that replaces manual label concatenation with encoder-derived representations. These are injected via AdaIN blocks placed before each upsampling layer, enabling progressive and localized domain modulation. Unlike fixed embeddings, our encoder processes reference inputs and dynamically infers bias features, allowing the model to learn domain transformations in a fully end-to-end fashion. This improves the flexibility of bias control and reduces reliance on explicit label priors. Full architectural details and training configurations are included in the released codebase.

We can also observe that even with the use of a standard StarGAN for bias translations, BLADE still performs better than the SOTA approach, i.e., CDvG, for the 0\% variant of data.

\section{Validating Bias Conflict Severity as a Metric}
\begin{table}[h]
\centering
\resizebox{0.7\linewidth}{!}{%
\begin{tabular}{c|cc|c}
& \multicolumn{2}{c|}{\textbf{Predicted}} & \\
\textbf{Actual} & Aligned & Conflicting & \textbf{Total} \\
\hline
Aligned & 52513 & 38 & 52551 \\
Conflicting & 452 & 1997 & 2449 \\
\hline
\textbf{Total} & 52965 & 2035 & 55000 \\
\end{tabular}
}
\caption{Confusion Matrix for Bias Classification on Colored MNIST}
\label{tab:confusion_matrix}
\end{table}
To prove the efficacy of bias conflict severity, we evaluate its ability to distinguish between bias-aligned and bias-conflicting samples on Colored MNIST 5\% BCR data. We define bias-aligned samples as those with a bias conflict severity below 0.5, while samples with a severity of 0.5 or higher are classified as bias-conflicting. As shown in Table~\ref{tab:confusion_matrix}, the confusion matrix illustrates the reliability of this threshold-based classification, effectively capturing the distinction between these two categories.

\end{document}